\documentclass{article}
\usepackage{arxiv}
\usepackage[utf8]{inputenc} 
\usepackage[T1]{fontenc}    
\usepackage{hyperref}       
\usepackage{url}            
\usepackage{booktabs}       
\usepackage{amsfonts}       
\usepackage{nicefrac}       
\usepackage{microtype}      
\usepackage{lipsum}		
\usepackage{graphicx}
\usepackage{doi}
\usepackage{algorithmic}
\usepackage{algorithm}
\usepackage{adjustbox}
\usepackage{amssymb}
\usepackage{subcaption}
\usepackage{amsmath}
\usepackage[T1]{fontenc}

\usepackage{multicol}
\usepackage{amssymb}
\usepackage{wrapfig}
\usepackage{amsthm}%
\usepackage{mathrsfs}%
\usepackage{xcolor}%
\usepackage{algorithm}%
\usepackage{algorithmic}
\usepackage{bm}
\usepackage{amsmath}
\usepackage{color}

\title{ECoDe: A Sample-Efficient Method for
Co-Design of Robotic Agents}

\author{ {\hspace{1mm}Kishan R. Nagiredla} \\
	Applied Artificial Intelligence Institute (A$^2$I$^2$)\\
	Deakin University, Australia \\
	\texttt{knagiredla@deakin.edu.au} \\
	\And
	{\hspace{1mm}Buddhika L. Semage} \\
	Applied Artificial Intelligence Institute (A$^2$I$^2$)\\
	Deakin University, Australia \\
 	\And
	{\hspace{1mm}Thommen G. Karimpanal} \\
	Applied Artificial Intelligence Institute (A$^2$I$^2$)\\
	Deakin University, Australia \\
 	\And
	{\hspace{1mm}Arun Kumar A. V} \\
	Applied Artificial Intelligence Institute (A$^2$I$^2$)\\
	Deakin University, Australia \\
    \And
	{\hspace{1mm}Santu Rana} \\
	Applied Artificial Intelligence Institute (A$^2$I$^2$)\\
	Deakin University, Australia \\
}

\date{}

\renewcommand{\shorttitle}{Sample-Efficient Co-Design of Robotic Agents Using Multi-fidelity Training \\ on Universal Policy Network}

\hypersetup{
pdftitle={A template for the arxiv style},
pdfsubject={q-bio.NC, q-bio.QM},
pdfauthor={David S.~Hippocampus, Elias D.~Striatum},
pdfkeywords={First keyword, Second keyword, More},
}

\begin{document}
\maketitle

\begin{abstract}
Co-designing autonomous robotic agents involves simultaneously optimizing the
controller and the agent’s physical design. Its inherent bi-level optimization formulation necessitates an outer loop design
optimization driven by an inner loop control optimization. This
can be challenging when the design space is large and each
design evaluation involves a data-intensive reinforcement learning
process for control optimization. To improve the sample efficiency of co-design,
we propose a multi-fidelity-based exploration strategy in
which we tie the controllers learned across the design spaces
through a universal policy learner for warm-starting subsequent
controller learning problems. Experiments performed on a wide
range of agent design problems demonstrate the superiority of
our method compared to baselines. Additionally, analysis
of the optimized designs shows interesting design alterations
including design simplifications and non-intuitive alterations.


\end{abstract}


\section{Introduction}\label{sec1}
Reinforcement Learning (RL) has been a prominent approach for training
agents to learn complex behaviors, relying solely on reward maximization.
Whilst most robotics
research is centered around a few well-known, fixed skeleton designs
e.g., robotic arms or bipedal humanoids, there is an abundance of skeleton designs in nature that equip animals with unique and powerful capabilities. For e.g., the split hoof design of Alpine Ibex makes them excellent climbers, or the strong hind legs make the Kangaroo rats the best jumpers, etc. Exploring such exotic design spaces can lead us to exceedingly more capable designs. Unfortunately, design optimization is a hard problem because the design space can be large, and evaluating designs can be computationally expensive, especially when the control is learned through inherently sample-intensive RL algorithms. 

A subset of the robot design problem that we consider in this work deals with fixed skeletal structures with variable parameters. 
Such problems are often formulated as bi-level optimization problems \cite{bhatia2021evolution}. This includes (a) searching over the design space in the outer loop and (b) evaluating each design's task-solving capability by learning the control policy in the inner loop. The inner loop typically involves training the agent with Deep-RL methods \cite{schulman2017proximal}, making it an extremely sample-intensive process. Ha \cite{Ha2019} used a Genetic Algorithm (GA) for optimization of the outer loop, and other notable works such as \cite{yuan2021transform2act}, learned a parameter-attribution policy using RL in the outer loop. Unfortunately, both GAs and RL are sample-intensive, and their combinations as such becomes practically infeasible. A nai\"ve solution may include running sample-efficient optimization algorithms such as Bayesian optimization \cite{pelikan1999boa} in the outer loop. However, such a solution would still be limited, as firstly, it ignores the fact that control policies corresponding to adjacent
regions in the parameter space may be similar to each other. As a result, the policy corresponding to each parameter would necessarily have to be learned from scratch, which can be highly inefficient. Secondly, such solutions ignore the stochastic monotonicity of RL; i.e., the fact that on average, the performance of RL agents tends to monotonically increase with time, which could form a basis prematurely terminating unpromising designs. 

To this end, we propose a novel approach in which the morphology and control policy required to perform a task are learned in conjunction. Our framework (a) employs transfer learning to exploit the closeness of the control policies corresponding to adjacent sets of parameters and (b) uses a type of multi-fidelity approach to exploit the stochastic monotonicity in RL. Specifically, we adopt the multi-armed bandit-based hyperparameter optimization method HyperBand \cite{Li2017}
that uses a set of multi-level filters, with each filter offering a
specific mix of exploration (how many different parameters are examined)
and exploitation (how well they are examined). In HyperBand, the widest filter starts with a large set of random parameters, where at the first level a fixed but small sampling budget is provided for evaluating each parameter. Larger sampling budgets are provided in subsequent levels, for evaluating a smaller set of more promising designs.

\begin{figure}
\begin{centering}
\includegraphics[scale=0.28]{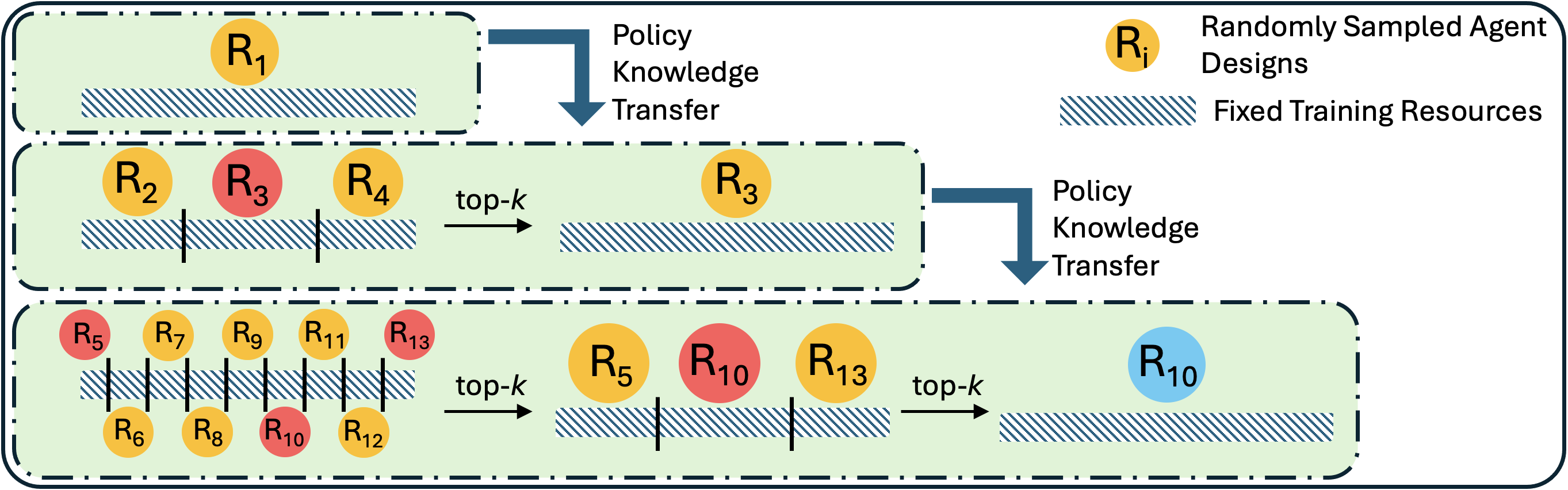}
\par\end{centering}
\begin{centering}
\caption{\label{fig:ecode} ECoDe Architecture for multi-fidelity based knowledge propagation mechanism to identify the best co-design (light blue blob). The blobs indicate different robot design samples and hatched boxes indicate the training time. Inside each horizontal box (light green), top-performing samples (red blobs) are made to progress from lower fidelities to higher fidelities. The thick blue arrows indicate the evaluation order to aid effective knowledge transfer through UPN.}
\par\end{centering}
\centering{}
\end{figure}

Since RL generally exhibits stochastic monotonicity, it is more likely that low-fidelity (the policy is trained with smaller number of samples) evaluations of policies are somewhat reflective of their high-fidelity (the policy trained with larger number of samples) evaluations. However, in HyperBand, since the ordering based on just the low-fidelity evaluation can be noisy, to improve the chances of retaining the best designs, a set of top-$k$ parameters are collected and trained with more samples. The whole process repeats until only one best-performing parameter remains. Subsequent filters start with lesser number of initial parameters, but this initial set is provided a larger budget in the first level than the preceding filter. This way HyperBand effectively uses stochastic monotonicity to efficiently navigate the sample space. Additionally, we use Universal Policy Network (UPN) \cite{Yu2020a} to perform transfer learning across parameters where the policy
learning for a new parameter borrows knowledge from the policy learned
with the last set of parameters. However, UPN-based transfer can create a biased preference for high-fidelity observations as even a bad design with a high number of samples to train its policy can create a seemingly better policy than a good design with only a small number of samples to train its policy. Thus, through Fig. \ref{fig:ecode} we show how ECoDe navigates the filters  in a specific manner (in the opposite direction to HyperBand - explained later in \ref{sec:methodolgy}) to reduce such bias.

We implement our method on seven OpenAI Gym \cite{brockman2016openai}
environments to demonstrate its effectiveness in identifying good
co-designs. We contend that these environments are well-suited for
our research, as they provide a standardized interface for training
RL algorithms across a wide range of environments and tasks, and closely mimic the type of control required for real-world robotics applications \cite{andrychowicz2020learning}. Our results show that our approach outperforms existing methods in identifying good co-designs across
multiple environments.

\section{Related Works}\label{sec2}

The co-design of physical and control structures for robotic agents \cite{yuan2021transform2act}  has for long, been a problem of interest.   
From Von Neumman\textquoteright s work around
the idea of agents utilizing evolutionary mechanisms \cite{VonNeumann1966} for morphing, similar mechanisms were 
later used to generate intelligent virtual agents \cite{Sims1994}. While some research continued to focus on the optimization of skeletal structures and design parameters, others focused
on using pre-selected skeletons to optimize solely over design
parameters. Nevertheless, early evolutionary mechanisms were extremely sample-inefficient \cite{Yu2020} as it involved evaluating each design in the population. Recent works by Ha \cite{Ha2019} and Yuan et al. \cite{yuan2021transform2act}  have addressed this with sample efficient approaches. Parts of the environment are parameterized, facilitating joint learning of policy and physical structure to uncover task-assistive design principles in both these works. By using the implicit function
theorem \cite{Jittorntrum1978}, Ha expresses both motion and design parameters
as functions describing robot dynamics. His work optimizes over
a linear approximation of these functions resulting in agents that
learn to change the policy and design parameters depending on
the task. 

Transfer learning has been used to improve sample efficiency in several works. Schaff et al. \cite{Schaff2019} demonstrated the transfer of knowledge between previous and new designs by considering design parameters as an additional policy input. Similarly, Luck et al. \cite{luck2020data} used the Q-function from RL as an objective function for design adaptation from randomly chosen initial designs. Model-based methods like Villarreal et al. \cite{VillarrealCervantes2012} rely on modeling environment dynamics to learn both morphological and control policies sample-efficiently. However, such methods lack robustness, and are sensitive to changes in the dynamics or design parameters, emphasizing the importance of generalization and knowledge transfer for co-design. In many domains, control policies modeled via neural networks
and trained using the Deep-RL framework have outperformed prior methods
and have delivered state-of-the-art results \cite{Zhao2020}, \cite{Yu2020a} \cite{semage2022fast}. Although their ability to learn complex policies without supervision is desirable, deep RL deals solely with policy optimization, and is as such, sample inefficient by nature. 

Recently Bayesian Optimization (BO) \cite{pelikan1999boa} and HyperBand \cite{Li2017} have successfully been used for sample efficient optimization, specifically in the context of hyperparameter optimization of deep neural networks.  While BO provides a general framework for black-box optimization, HyperBand provides a more specific solution that exploits stochastic monotonicity to achieve further sample efficiency through multi-fidelity search. Since RL also demonstrates stochastic monotonicity with respect to the number of steps, HyperBand provides a more natural choice to work in conjunction with RL techniques. In this work, we focus on improving the sample efficiency of co-design by leveraging multi-fidelity methods and the knowledge-sharing aspect of existing policy transfer mechanisms. 

\section{Background}\label{sec3}

\subsection{Reinforcement Learning}\label{subsec3}
Reinforcement Learning (RL) problems are modeled as a Markov Decision
Process (MDP), represented as $\langle\mathcal{S},\mathcal{A},T,R\rangle$, where
$\mathcal{S}$ is the state-space, $\mathcal{A}$ is the action-space,
$T:\mathbb{\mathcal{S}\times\mathcal{A}\rightarrow\mathcal{S}}$ is
the transition function governing the next state reached by taking
an action $a\in\mathcal{A}$ in a state $s\in\mathcal{S}$, and $R:\mathcal{S}\times\mathcal{A}\rightarrow\mathbb{R}$
is the scalar-valued reward-producing function for taking action $a$ in state
$s$. 

The learning problem is to find the optimal policy that maximizes the returns $\mathbb{E}_{\tau\sim\pi}R(\tau)$, where $\pi$ is the policy, $\tau$ is a trajectory sampled from $\pi$, and $R(\tau)=\sum_{(s,a)\in\tau}R(s,a)$ is shorthand for sum of the rewards over the trajectory $\tau$.

\subsection{\label{subsec:Universal-Policy-Network}Universal Policy Network
(UPN)}

UPN \cite{Yu2020a} is an RL approach to simultaneously learn a library of policies
corresponding to different design parameters. UPN state $s_{\text{UPN}}\in\mathcal{S}_{\text{UPN}}$ is obtained by augmenting the agent's state $s$ with design parameters $\theta$ as $s_{\text{UPN}} = [s, \theta]^\intercal$. 

UPN policy $\pi^{UPN}$ is learned by solving the MDP  $\langle\mathcal{S}_{UPN},\mathcal{A},T_{UPN},R\rangle$, where $T_{UPN}:S_{UPN}\times \mathcal{A}\rightarrow S_{UPN}$.
Assuming that the control problems between two close sets of design parameters are not very different, $\pi^{UPN}$ can also be differentiable and thus learnable. UPN has been shown to exploit this property \cite{Yu2020a} by learning across different design parameters together via large neural networks. 

\subsection{\label{subsec:Hyperband}HyperBand}

HyperBand \cite{Li2017} is a powerful bandit-based multi-fidelity technique that extends the capability and generalizability of a much simpler but effective technique - Successive Halving \cite{Jamieson2016}, designed to stop poorly performing configurations early. HyperBand shows a remarkable performance boost in solving the problem of Hyperparameter Optimization in deep neural networks and outperforms random search and Bayesian Optimization \cite{Li2017}.

\begin{table}
\centering{}%
\caption{\label{tab:Hyperband-process-showcasing}HyperBand with 4 filters ($F$) and their corresponding number of configurations ($n_{i}$) and  resource units ($p_{i}$) when $M$=27 and $\eta$=3.}
\vspace{0.2cm}
\begin{tabular}{|c|cc|cc|cc|cc|}
\hline 
 & \multicolumn{2}{c|}{$F=3$} & \multicolumn{2}{c|}{$F=2$} & \multicolumn{2}{c|}{$F=1$} & \multicolumn{2}{c|}{$F=0$}\\
$i$ & $n_{i}$ & $p_{i}$ & $n_{i}$ & $p_{i}$ & $n_{i}$ & $p_{i}$ & $n_{i}$ & $p_{i}$ \\
\hline 
0 & 27 & 1 & 9 & 3 & 3 & 9 & 1 & 27\\
1 & 9 & 3 & 3 & 9 & 1 & 27 &  & \\
2 & 3 & 9 & 1 & 27 &  &  &  & \\
3 & 1 & 27 &  &  &  &  &  & \\
\hline 
\end{tabular}
\end{table}

In HyperBand, a given budget $B$ is partitioned into a combination of a number of configurations $(M)$.  A budget of $(F_{max}+1)M$ is allocated per configuration in each filter, where $F_{max}$ is the maximum number of filters (obtained from { $\left\lfloor log_{\eta}(M)\right\rfloor $}). These filters (arranged from highest to lowest exploration as presented in Table \ref{tab:Hyperband-process-showcasing}) are independent and resemble arms in a multi-armed bandit
technique. Successive Halving is then called as a subroutine on the randomly sampled configurations $n_{i}$ and a resource budget of $p_{i}$ is allocated, which then outputs the top-$k$ performers.  

\section{Methodology}\label{sec:methodolgy}

We aim to determine our optimal design parameters $\theta^{*}$, given by:

\begin{equation}
    {\theta}^{*} = \arg\max_{{\theta} \in \Theta} f({\theta})
\end{equation}
where ${\theta}$ represents the design parameters to be learned, $f()$ is a performance measure for a particular design choice, and $\Theta$ is the space of choices for the design parameters, which we assume to be a bounded set.

In many cases $f()$ can be time-consuming to measure, as learning a policy for complex control problems will require a large number of observations ($\{s,a,R(s,a)\}$). Instead, we consider obtaining a low-fidelity (noisy) evaluation of the policy by providing it with a limited budget of resources for evaluation. The key idea is that although low-fidelity evaluations may be noisy, they may still be reflective of the quality of the designs. For instance, it may be possible to use low-fidelity measurements to discard low-performing designs and reserve the high-fidelity evaluations for more promising design configurations. This manner of multi-fidelity evaluations boosts sample efficiency by ensuring that a large number of samples is not spent for accurately evaluating clearly inferior design configurations. Such problems of optimization of expensive functions, exploiting the multi-fidelity problem is common in neural network HyperParameter Optimization problems (HPO) where $\mathbf{\theta}$ would be the set of hyperparameters to be tuned, $f()$ being the validation performance.
Even in such cases, multi-fidelity measurement of $f()$ can be achieved by limiting the number of training epochs. One prominent HPO algorithm in that context is HyperBand, discussed in \ref{subsec:Hyperband}.

Though HyperBand has been proven to show good performance, adaptability as well as scalability in high-dimensional spaces, it does not attempt to learn across configurations. For instance, once a design configuration is evaluated using the multi-fidelity approach described above, one would need to redo the evaluation for another configuration, even if it is highly similar to the former. We address this by using UPN, which trains the required configuration while simultaneously updating the policy parameters of neighboring configurations. As a result, when applying the multi-fidelity approach to neighboring configurations, we obtain a much more reliable estimate of its performance.  

In the original HyperBand (Table \ref{tab:Hyperband-process-showcasing}), filtering is done from left to right ($F=3$ to $F=0$), starting with low-fidelity filters, and subsequently moving to high-fidelity ones. Since the nature of low-fidelity filters is to obtain noisy evaluations of several configurations, and that of high-fidelity filters is to obtain more reliable evaluations of specific configurations, using UPN in this same direction may not be the best choice. Through ECoDe, we propose that UPN if instead was used in reverse (right to left), would start with high-fidelity evaluations on a smaller number of configurations, using which policies corresponding to multiple configurations would be updated (as UPN tends to train policies for multiple configurations simultaneously). As a result, the evaluations of more configurations in subsequent lower fidelity filters would be more reliable, thereby further improving the sample efficiency of the co-design process. 
 
To summarize, ECoDe evaluates and filters out ineffective co-designs
and continually learns from successful co-designs to achieve good task performance in a sample-efficient manner. Our proposed method is provided in Alg. \ref{alg:ECoDe}. 

\begin{algorithm}
\begin{algorithmic}[1]
\STATE \textbf{Input}: $M$ (maximum configurations), $\eta$ (early-stopping aggressiveness)
\STATE \textbf{Initialize} $F_{max}=\left\lfloor log_{\eta}(M)\right\rfloor, B=(F_{max}+1)M$ 

\STATE //$F_{max}+1$: maximum number of filters, $B$: budget per filter

\FOR{  $F \in\ 0,1,.....,F_{max}-1,F_{max}$}

    \STATE //Begin Successive\_Halving by calculating $n$ and $p$ values. 
    \STATE { $n=\left\lceil \frac{B}{M}\frac{\eta^{F}} {(F+1)}\right\rceil $}$\quad$  //$n$: configurations per filter
    \STATE { $p=M\eta^{-F}$}$\quad\quad$ //$p$: the resource units per filter
    \STATE //Collect $n$ different hyperparameter (design) configurations in $T$
    \STATE { $T=get\_hyperparameter\_configuration(n)$}
    \FOR{ $i\in\{0,.....,F\}$ }
        \STATE $n_{i}=\left\lfloor n\eta^{-i}\right\rfloor$ \quad //$n_{i}$: configurations per round
        \STATE $p_{i}=p\eta^{i}$ \qquad //$p_{i}$: resource units per round 
        \STATE {$L=\{run\_UPN(t,p_{i}):t\in T\}$} //Call run\_UPN() to store average rewards collected by each agent $t$ in $T$.
        \STATE {Based on $\eta$ value, discard low-performers and update $T$}
        \STATE $T=$ top-$k(T,L,\left\lfloor n_{i}/\eta\right\rfloor$) //Using $\eta$ discard low-performers and update $T$. 
    \ENDFOR    
\ENDFOR
\STATE \textbf{return} $t = max(T)$
\caption{ECoDe (Efficient Co-Design)}
\label{alg:ECoDe}
\end{algorithmic}
\end{algorithm}

\section{Experiments and Results}

\begin{figure}
\begin{centering}
\includegraphics[scale=0.6]{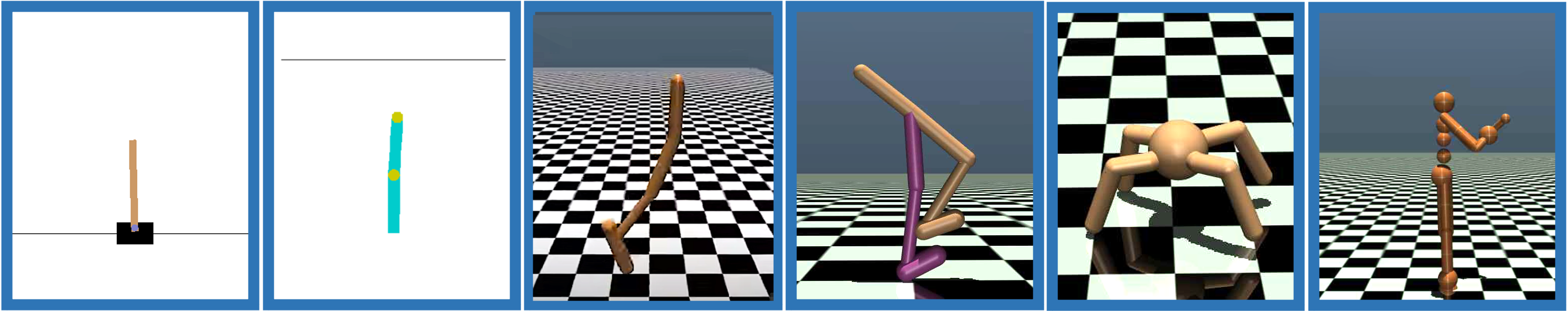}
\par\end{centering}
\begin{centering}
\caption{\label{fig:allenvs} OpenAI Gym Environments (left to right) - CartPole, Acrobot, Hopper, Walker2D, Ant and Humanoid.}
\par\end{centering}
\centering{}
\end{figure}
\subsection{\label{subsec:Environments}Environments}
Our simulation environments consist of 2 modified Classic control and 5 modified Mujoco \cite{todorov2012mujoco} environments
from OpenAI Gym \cite{brockman2016openai} (shown in Fig. \ref{fig:allenvs}). Specifically, the goal of all Mujoco agents is to move forward without falling to collect the maximum reward. 

\noindent \textbf{CartPole (Classic)}: We use the standard CartPole-v1 environment which has a cart balancing a pole of a given length, and allow for the pole length to be changed in the range $[0.1, 3]$ to maintain the pole upright. Additionally, we modified the reward function to impose a ground truth length of 1.425.

\noindent \textbf{Acrobot (Classic): } We use the Acrobot-v1 environment, where the agent's goal is to provide enough actuation at the intersection of the two links to get the second
link to reach the target height. The configurable design parameters
are the link lengths and their masses, which vary between 0.1 and 2.

\noindent \textbf{Hopper (Mujoco):} We use the Hopper-v3 environment in which the 2D one-limb robot has a torso, thigh, leg, and foot. The Foot and Leg lengths are allowed to vary
between 1/4th and 4 times their original lengths.

\noindent \textbf{Walker2D (Mujoco):} We use the Walker2D-v3 environment in which the two-limb robot contains 4 main parts: a torso, two thighs,
two legs, and two feet. All these parts are allowed to vary within $\pm20\%$ of their original values.

\noindent \textbf{Ant (Mujoco):} We use the standard Ant-v3 environment in which the agent is a 3D ant-like four-limb robot with a spherical torso. The design parameters are the link lengths
of all 4 limbs. The range is chosen to be between 0.1 to 0.5.

\noindent \textbf{Humanoid (Mujoco):} We use the standard Humanoid-v3 environment in which the agent is a 3D human-like two-legged and two-armed robot. We have considered thigh, shin, and feet length of both leg lengths as the design parameters that can vary between $0.5$ to $1.5$ times the default values.

\noindent \textbf{MaxHumanoid (Mujoco):} This environment is similar to the original Humanoid except that we allow for 16 configurable parameters. This presents a highly complex co-design problem where the agent has to choose length and thickness values for all 12 parts in the four limbs with a range similar to that of the original Humanoid environment (0.5 to 1.5 times the default values).

\subsection{Baselines}

We evaluate our method against the following baselines:

\noindent \textbf{RandomSearch:} We randomly sample parameters
from a uniform distribution within their ranges. This provides a lower bound on the expected performance of other methods.

\noindent \textbf{Transform2Act \cite{yuan2021transform2act}:} Transform2act uses graph-based
representation for agents with limbs represented as edges and joints
as nodes in a 3-stage policy optimization - first for skeleton
design, second for parameter learning, and then for learning
control policy. In our implementation of Transform2Act, we freeze
Transform2act's first stage as we assume the skeleton to remain the same.

\noindent \textbf{nLimb \cite{Schaff2019}:} nLimb uses a Gaussian mixture model to parameterize the design distribution and maintains multiple different hypotheses for designs that may be promising. Their method maintains a uniform distribution across components, and half of them with the lowest rewards are eliminated after every $N$ iterations.

\noindent \textbf{HyperBand \cite{Li2017}:} In original form, without UPN.  

\noindent \textbf{Coadaptation \cite{luck2020data}:} This approach utilizes the Q-function of a trained policy to evaluate the suitability of given design parameters. Subsequently, it employs particle swarm optimization in conjunction with exploration heuristics to identify the next viable design parameters for evaluation.

We compare these methods with our proposed method 

(discussed in \ref{sec:methodolgy})
to make the co-optimization sample efficient. All methods
are given a budget of a fixed number of steps that can be taken in an environment. Our code is available at \url{https://n-kish.github.io/ecode/}.  

\subsection{UPN Framework}
The UPN agent architecture we use is a dense network with 3 hidden layers, each containing 64 nodes. The input for the network is the concatenated vector of observations (o) for the task with design parameters ($\theta$) for the agent and the environment. The output of this network produces a policy distribution (mean and std. variation) with the size of action space for the environment.

\subsection{Policy Performance}

Table \ref{tab:Summary-Analysis-of-all} shows the average rewards accumulated
by the best parameter-policy combinations for different environments
and by different algorithms.
We used a budget of 2.13 million steps for CartPole, Acrobot, Hopper, and Walker2D and an increased 2.84 million steps for Ant and 7.11 million steps for Humanoid, as
Ant and Humanoid are more complex control problems. 

\begin{table}
\caption{\label{tab:Summary-Analysis-of-all} Average Performance across 10 different runs of the best
design-policy combination across various environments and
algorithms. All algorithms have been given the same sampling budget
(2.13 million for all, except for the Ant, which uses 2.84 million
steps, and Humanoid, which uses 7.11 million steps).}
\begin{adjustbox}{width=\textwidth}
\begin{tabular}{|c|c|c|c|c|c|c|}
\hline 
 & \multicolumn{1}{c|}{{CartPole}} & {Acrobot} & {Hopper} & {Walker2D} & {Ant} & {Humanoid}\tabularnewline
\hline 
{Random Search} & {328.4 $\pm$ 4.7} & {-434.6 $\pm$ 2.9} & {587.1 $\pm$ 21.2} & {138.3 $\pm$ 3.7} & {-1183.1 $\pm$ 22.1} & {774.1 $\pm$ 28.8}\tabularnewline 
{HyperBand \cite{Li2017}} & {432.3 $\pm$ 9.1} & {-33.2 $\pm$ 7.6} & {823.8 $\pm$ 28.4} & {428.7 $\pm$ 4.1} & {-33.4 $\pm$ 2.7} & {856.8 $\pm$ 18.8}\tabularnewline
{Transform2Act \cite{yuan2021transform2act}} & {-} & {-} & {545.1 $\pm$ 2.8} & {614.1 $\pm$ 4.3} & {533.3 $\pm$ 9.3} & {613.2 $\pm$ 8.1}\tabularnewline

{nLimb \cite{Schaff2019}} & {-} & {-} & {967.3 $\pm$ 5.4} & {1344.5 $\pm$ 6.6} & {691.9 $\pm$ 5.7} & {645.5 $\pm$ 3.2}\tabularnewline

{Coadaptation \cite{luck2020data}} & {444.4 $\pm$ 5.9} & {-29.5 $\pm$ 2.8} & {867.5 $\pm$ 71.4} & {1568.9 $\pm$ 198.3} & {1018.3 $\pm$ 203.9} & {819.4 $\pm$ 26.9}\tabularnewline
{ECoDe} & \textbf{464.1 $\pm$ 7.9} & \textbf{-12.9 $\pm$ 2.4} & \textbf{1089.4 $\pm$ 7.1} & \textbf{3297.4 $\pm$ 179.6} & \textbf{3419.1 $\pm$ 128.2} & \textbf{{}4110.3 $\pm$ 179.5}\tabularnewline
\hline 
\end{tabular}
\centering{}
\end{adjustbox}
\end{table}

The results in Table \ref{tab:Summary-Analysis-of-all} represent an average of 10 independent trials along with the standard errors.  We implemented Transform2Act and nLimb on the Hopper, Walker2D, and Ant environments and we extended their application to the Humanoid environment. We implemented Coadaptation from scratch and hence were able to compare it with ECoDe on all the environments. Except for CartPole environment, ECoDe significantly outperforms the second-best (i.e. with significance level, p $<$ 0.05) in all other environments. In CartPole, ECoDe is also better than the second-best but at a slightly reduced significance level (i.e. with p = 0.06). This shows the utility of knowledge propagation
via UPN and the efficient use of environment interactions through multi-fidelity evaluations. 

Surprisingly, Transform2Act sometimes performed worse
than Random Search, possibly due to it trying to solve a much
harder problem (i.e. learning a policy to choose good designs,
rather than directly learning a good design). In Hopper, Humanoid \& MaxHumanoid, the default reward function is such that
the agents move forward while balancing themselves with minimal control maneuvers. ECoDe seems
to focus on getting the balance first by getting the physical
design parameters in the right zone (e.g., increase the foot
size) and then choosing the best parameter by finding the one
easiest to control. This intuitive breakdown of the
design process is akin to what an expert might have done.  

\subsection{Design Search Optimality}
The results in Table \ref{tab:Summary-Analysis-of-all} showcase ECoDe's ability to find co-designs that perform significantly better compared to other algorithms across multiple environments.
Table \ref{tab:Skeleton-analysis-cartpole} shows the corresponding co-design pole lengths
for the CartPole environment as found by different algorithms when
given different amounts of sampling budget As seen, even across
different sampling budgets, ECoDe found designs,
which are closest to the ground truth optimal length of 1.425.

\begin{table}
\caption{\label{tab:Skeleton-analysis-cartpole}Design analysis (as mean $\pm$
standard error) of Random Search, HyperBand and ECoDe algorithms in CartPole Environment over 10 trials
when trained with a budget of 710K steps, 2.13M steps and 2.84M steps.}
\begin{centering}
\begin{tabular}{|c|c|c|c|}
\hline 
 & \multicolumn{3}{c|}{{Pole Lengths per Sampling Budgets}}\tabularnewline 
 \hline
 {Methods}& {710K} & {2.13M} & {2.84M}\tabularnewline
\hline 
{RandomSearch} & {1.51$\pm$ 0.1} & {1.55 $\pm$ 0.1} & {1.86 $\pm$ 0.1}\tabularnewline
{HyperBand} & {1.11 $\pm$ 0.2} & {1.17 $\pm$ 0.1} & {1.11 $\pm$ 0.2}\tabularnewline
{ECoDe} & \textbf{1.45 $\pm$ 0.1} & \textbf{1.5 $\pm$ 0.1} & \textbf{1.34 $\pm$ 0.1}\tabularnewline
\hline
\end{tabular}{\par}
\par\end{centering}
\centering{}
\end{table}

With abundant samples and a powerful network that can learn
intricate control rules, the length selection seems less important
for achieving higher rewards. However, ECoDe still provides an optimal length closest to the ground truth.

\subsection{Computational Efficiency}
The total number of agent-environment interactions (the most time-consuming aspect in RL) for ECoDe is $O(\omega nlogn)$ where $n$ is the number of different configurations that we must try and $\omega$ is the minimum number of agent-environment interactions that must be allowed. Thus, we see that it uses the total resources quite efficiently. Table \ref{tab:runtime} shows the wall-clock time for each environment when run on a Dell R6525 2.25GHz CPU with 64 cores and 1TB RAM machine.  

\begin{table*}[h]
\caption{\label{tab:runtime} Total runtime ECoDe needs to solve each of the OpenAI Gym environment\'s tasks. The runtime is dependent on the complexity of the agent and the environment and hence increases from CartPole to MaxHumanoid.}
\centering
\begin{adjustbox}{width=\textwidth}
\begin{tabular}{|c|c|c|c|c|c|c|c|}
\hline 
& {CartPole} & {Acrobot} & {Hopper} & {Walker2D} & {Ant} & {Humanoid} & {MaxHumanoid}\tabularnewline
 \hline 
{Runtime (in hours)} & {0:35} & {0:45} & {1:30} & {1:30} & {6:45} & {8:30} & {9:00} \tabularnewline
\hline
\end{tabular}
\centering{}
\end{adjustbox}
\end{table*}
Amongst all the baselines, Transform2Act, nLimb, Coadaptation, and ECoDe collected much higher average rewards across multiple environments, possibly owing to the policy-transfer mechanisms associated with them. However, while ECoDe took 1.5 hours to train the bipedal walker in the Walker2D environment, Transform2Act required 6 hours, and the average performance of their co-design is still sub-par (Table \ref{tab:Summary-Analysis-of-all}). We observe a similar trend in Hopper and Ant environments as well. In contrast, nLimb and Coadaptation matched our computational efficiency, although ECoDe's multi-fidelity based resource allocation mechanism facilitated collecting high average rewards in similar time.

\subsection{Design Simplification}

\begin{figure}
     \centering
     \begin{subfigure}[b]{0.4\textwidth}
         \centering
         \includegraphics[scale=0.2]{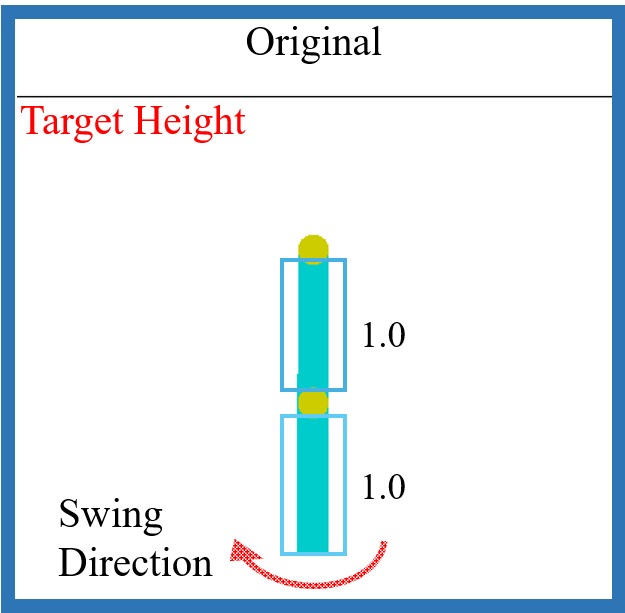}
         \label{fig:acro_l}
     \end{subfigure}
     \hspace{0.3cm}
     \begin{subfigure}[b]{0.4\textwidth}
         \centering
         \includegraphics[scale=0.2]{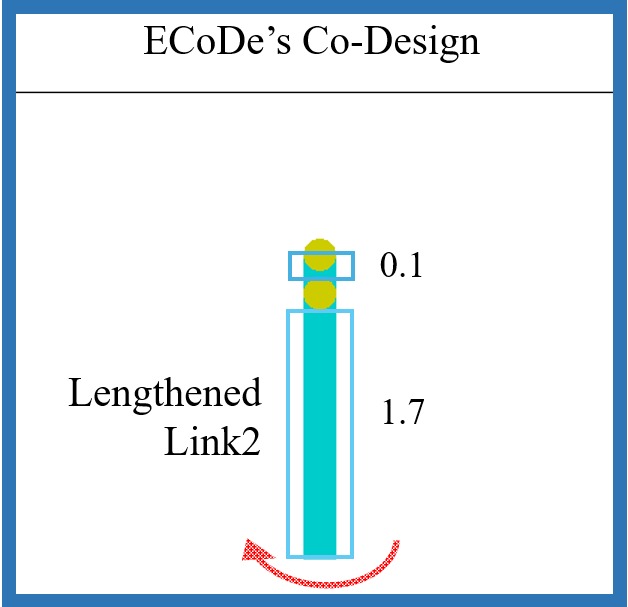}
         \label{fig:acro_r}
     \end{subfigure}
        \caption{The original Acrobot (left) vs. ECoDe simplified Acrobot (right). The simplified version resembles a pendulum, making the control problem easier.}
        \label{fig:Acrobot-visualisation}
\vspace{-0.3cm}
\end{figure}

A rather more interesting result is observed in the Acrobot environment. The best-performing co-design configuration happens to find
a design that reduced the more difficult two-link Acrobot control
problem to a simpler one-link control problem by choosing the smallest
length value for the first link (Fig. \ref{fig:Acrobot-visualisation}).

\subsection{Constrained Design}
\begin{figure}
     \centering
     \begin{subfigure}[b]{0.4\textwidth}
         \centering
         \includegraphics[scale=0.4]{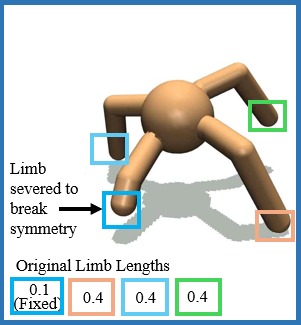}
         \label{fig:ant_l}
     \end{subfigure}
     \hspace{0.3cm}
     \begin{subfigure}[b]{0.4\textwidth}
         \centering
         \includegraphics[scale=0.4]{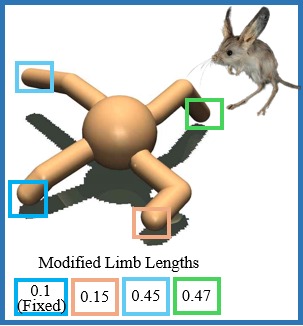}
         \label{fig:ant_r}
     \end{subfigure}
        \caption{Ant robot with a broken front left limb (left, blue box) and the ECoDe-suggested co-design (right) with a shortened front right limb (orange box) and lengthened hind limbs (yellow and grey boxes), resembling a kangaroo rat.}
        \label{fig:antECoDeR-figures}
\vspace{-0.7cm}
\end{figure}

We experimented with breaking the robot's symmetry in
the Ant environment by removing the front left leg's ground contact link (Fig. \ref{fig:antECoDeR-figures}). While
other methods struggled to identify designs that walk, our method found good co-designs within a 2.8 million steps budget. The resultant design had the other front leg (i.e., right leg) shortened and the hind legs elongated, resembling a Kangaroo rat.
This bio-mimetic design exhibits jumping and crawling behaviours, as shown in supplementary videos at \url{https://n-kish.github.io/ecode/}. Although the control policy can be further refined, the design appears near-optimal given the limited time steps.

\subsection{Scalability}

We use the MaxHumanoid Environment discussed in  \ref{subsec:Environments} to create a humanoid with 16 configurable design parameters. Remarkably, with the same budget of 10.2 million timesteps, our agent was able to identify the best co-design and learn the complex task of humanoid walking.

\section{Conclusion}

We presented ECoDe, a multi-fidelity-based co-design method that uses a transfer learning mechanism to efficiently discover optimal design-control policy combinations for robotic agent design. Specifically, we perform a multi-fidelity search whilst warm-starting policies so that inferior designs can be identified and discarded using fewer samples, resulting in a sample-efficient co-design method. We evaluated our method on 7 different robot design problems using realistic physics simulators and the results show that on all occasions ECoDe performed the best. An interesting future direction may include going beyond just design parameters and adapting ECoDe to co-design the skeletal structure as a whole.     

higher value of $M$ means ECoDe can draw more random independent design configurations to learn from and is thus tuned based on the complexity of the control problem and environment.

\bibliographystyle{unsrt}  
\bibliography{references}

\begin{thebibliography}{10}

\bibitem{bhatia2021evolution}
Jagdeep Bhatia, Holly Jackson, Yunsheng Tian, Jie Xu, and Wojciech Matusik.
\newblock Evolution gym: A large-scale benchmark for evolving soft robots.
\newblock {\em In Neural Information Processing Systems}, 34:2201--2214, 2021.

\bibitem{schulman2017proximal}
John Schulman, Filip Wolski, Prafulla Dhariwal, Alec Radford, and Oleg Klimov.
\newblock Proximal policy optimization algorithms.
\newblock {\em arXiv preprint arXiv:1707.06347}, 2017.

\bibitem{Ha2019}
David Ha.
\newblock Reinforcement learning for improving agent design.
\newblock {\em Artificial life}, 25(4):352--365, 2019.

\bibitem{yuan2021transform2act}
Ye~Yuan, Yuda Song, Zhengyi Luo, Wen Sun, and Kris~M Kitani.
\newblock Transform2act: Learning a transform-and-control policy for efficient agent design.
\newblock In {\em International Conference on Learning Representations}, 2021.

\bibitem{pelikan1999boa}
Martin Pelikan, David~E Goldberg, Erick Cant{\'u}-Paz, et~al.
\newblock Boa: The bayesian optimization algorithm.
\newblock In {\em Proceedings of the genetic and evolutionary computation conference GECCO-99}, volume~1, 1999.

\bibitem{Li2017}
Lisha Li, Kevin Jamieson, Giulia DeSalvo, Afshin Rostamizadeh, and Ameet Talwalkar.
\newblock Hyperband: A novel bandit-based approach to hyperparameter optimization.
\newblock {\em The Journal of Machine Learning Research}, 18(1), 2017.

\bibitem{Yu2020a}
Wenhao Yu, Jie Tan, Yunfei Bai, Erwin Coumans, and Sehoon Ha.
\newblock Learning fast adaptation with meta strategy optimization.
\newblock {\em IEEE Robotics and Automation Letters}, 5(2), 2020.

\bibitem{brockman2016openai}
Greg Brockman, Vicki Cheung, Ludwig Pettersson, Jonas Schneider, John Schulman, Jie Tang, and Wojciech Zaremba.
\newblock Openai gym.
\newblock {\em arXiv e-prints}, 2016.

\bibitem{andrychowicz2020learning}
OpenAI:~Marcin Andrychowicz, Bowen Baker, Maciek Chociej, Rafal Jozefowicz, Bob McGrew, Jakub Pachocki, Arthur Petron, Matthias Plappert, Glenn Powell, Alex Ray, et~al.
\newblock Learning dexterous in-hand manipulation.
\newblock {\em International Journal of Robotics Research}, 2020.

\bibitem{VonNeumann1966}
John Von~Neumann and Arthur~W Burks.
\newblock Theory of self-reproducing automata.
\newblock {\em IEEE Transactions on Neural Networks}, 5(1), 1996.

\bibitem{Sims1994}
Karl Sims.
\newblock Evolving virtual creatures.
\newblock In {\em Proceedings of the 21st annual conference on Computer graphics and interactive techniques}, 1994.

\bibitem{Yu2020}
Tong Yu and Hong Zhu.
\newblock Hyper-parameter {O}ptimization: A {R}eview of {A}lgorithms and {A}pplications.
\newblock {\em arXiv preprint arXiv:2003.05689}, 2020.

\bibitem{Jittorntrum1978}
K~Jittorntrum.
\newblock An implicit function theorem.
\newblock {\em Journal of Optimization Theory and Applications}, 25(4), 1978.

\bibitem{Schaff2019}
Charles Schaff, David Yunis, Ayan Chakrabarti, and Matthew~R Walter.
\newblock Jointly learning to construct and control agents using deep reinforcement learning.
\newblock In {\em 2019 International Conference on Robotics and Automation (ICRA)}, pages 9798--9805. IEEE, 2019.

\bibitem{luck2020data}
Kevin~Sebastian Luck, Heni~Ben Amor, and Roberto Calandra.
\newblock Data-efficient co-adaptation of morphology and behaviour with deep reinforcement learning.
\newblock In {\em Conference on Robot Learning}. PMLR, 2020.

\bibitem{VillarrealCervantes2012}
Miguel~G Villarreal-Cervantes, Carlos~A Cruz-Villar, Jaime Alvarez-Gallegos, and Edgar~A Portilla-Flores.
\newblock Robust structure-control design approach for mechatronic systems.
\newblock {\em IEEE/ASME Transactions on Mechatronics}, 18(5):1592--1601, 2012.

\bibitem{Zhao2020}
Wenshuai Zhao, Jorge~Pe{\~n}a Queralta, and Tomi Westerlund.
\newblock Sim-to-real transfer in deep reinforcement learning for robotics: a survey.
\newblock In {\em 2020 IEEE symposium series on computational intelligence (SSCI)}, pages 737--744, 2020.

\bibitem{semage2022fast}
Buddhika~Laknath Semage, Thommen~George Karimpanal, Santu Rana, and Svetha Venkatesh.
\newblock Fast model-based policy search for universal policy networks.
\newblock In {\em 2022 26th International Conference on Pattern Recognition (ICPR)}. IEEE, 2022.

\bibitem{Jamieson2016}
Kevin Jamieson and Ameet Talwalkar.
\newblock Non-stochastic best arm identification and hyperparameter optimization.
\newblock In {\em Artificial intelligence and statistics}, 2016.

\bibitem{todorov2012mujoco}
Emanuel Todorov, Tom Erez, and Yuval Tassa.
\newblock Mujoco: A physics engine for model-based control.
\newblock In {\em IEEE/RSJ international conference on intelligent robots and systems}, 2012.

\end{thebibliography}

\end{document}